\documentclass[letterpaper]{article} 
\usepackage{aaai2027}  
\copyrighttext{arXiv preprint}
\usepackage[hyphens]{url}  
\usepackage{graphicx} 
\usepackage{natbib}  
\usepackage{caption} 
\usepackage{algorithm}
\usepackage{algorithmic}
\usepackage{amsmath}
\usepackage{amssymb}
\usepackage{tabularx}
\newcommand{\rev}[1]{#1}

\usepackage{listings}
\DeclareCaptionStyle{ruled}{labelfont=normalfont,labelsep=colon,strut=off} 
\floatstyle{ruled}
\newfloat{listing}{tb}{lst}{}
\floatname{listing}{Listing}

\usepackage{booktabs}

\usepackage{xcolor} 

\title{The Dual-Tier Sentiment Framework: Scalable Multimodal UX Attribution in Web Search}

\author{
    Yi Ding\thanks{Corresponding author.},
    Gaurav~Kumar, 
    Rui~Min, 
    Isabella~Ye, 
    Bhavika~Goyal, 
    Emilio~Antunez,
    Huihui~Chen,
    Ke~Xu, 
    Xiaowei Sun, 
    Xin Xu, 
    Huijie Feng, 
    Pradyumna Narayana, 
    Golnaz Farhadi
}
\affiliations{
    Google, Mountain View, CA, USA \\
    ellading@google.com
}

\begin{document}

\maketitle

\begin{abstract}
Sustaining user experience (UX) quality in large-scale web search requires scalable evaluation. Traditional surveys are slow and bottleneck rapid A/B testing, while automated metrics often lack the granularity to diagnose visual-spatial UX anomalies. To address this, we introduce the Dual-Tier Sentiment Framework, a novel industrial architecture for automated UX attribution. Tier 1 (Macro System) acts as a scalable user sentiment regression detector, utilizing a fine-tuned encoder-decoder LLM (based on Gemini XS) on telemetry features. Through multi-task regularization and contextual feature engineering, this tier achieves a +9.8\% PR-AUC gain on over 500,000 held-out records, matching baseline recall on all validated regressions with around 20\% relative precision gain among the 12 held-out experiments. In production, it reduces measurement latency from weeks to hours and cuts survey overhead by over 40\%. Tier 2 (Micro System) provides granular diagnostics by employing a Multimodal LLM (Gemini 3.0 Pro) to natively assess visual renderings and spatial layout of web result blocks, quantifying ``User Toil'' as the estimated effort required to reach a satisfactory result. Enhanced by RL-based automated prompt optimization and an AI judge for win/loss patterns, the MLLM yields a >6.5\% F1 score improvement, offering deep insights into content quality and layout friction. Ultimately, this framework enables scalable, accurate, and diagnostic UX attribution, processing O(100k) live experiment queries daily.
\end{abstract}


\section{Introduction}


The continuous evolution of modern web search and advertising platforms is driven by massive, high-frequency A/B experiments \citep{Kohavi_Tang_Xu_2020}. While these rapid release cycles are essential for product innovation, accurately capturing how users subjectively experience these changes at scale remains a critical industry challenge. Exposing users to localized layout friction or visual degradation erodes platform trust and diminishes long-term engagement. Consequently, developing sophisticated, automated systems to evaluate User Experience (UX) is essential.

A central difficulty in this domain is the challenge of scalable Multi-modal UX Attribution. In a production environment, evaluating search satisfaction presents distinct hurdles compared to traditional offline, text-based relevance scoring:

\begin{itemize}
    \item \textbf{The Latency and Scalability Bottleneck:} Human-in-the-loop surveys provide qualitative insights but are fundamentally unscalable. Relying on human annotators delays the assessment of live experiments by weeks and is susceptible to inherent rater noise and response bias \citep{karpinska2021perils}.

    \item \textbf{The Diagnostic Gap in Aggregated Telemetry:} Automated telemetry (e.g., clicks) provides necessary global safeguards but lacks the granularity to explain \textit{why} a regression occurred. Standard logs struggle to detect visual-spatial UX anomalies that result in user friction in the subjective browsing journey. Legacy heuristic DOM-based baselines (e.g., element density or scroll depth) are similarly insufficient because they fail to capture \textit{semantic} visual friction, such as distinguishing an ad that is structurally identical in the DOM yet visually distinct (e.g., featuring a subtle `Sponsored' badge) from an organic result.

    \item \textbf{Multimodal Complexity:} Modern Search Result Pages (SRPs) are complex visual structures. Assessing user sentiment requires bridging sparse behavioral logs with the multimodal nature of the rendered page structure.
\end{itemize}

\begin{figure*}[t]
    \centering
    \includegraphics[width=0.95\textwidth]{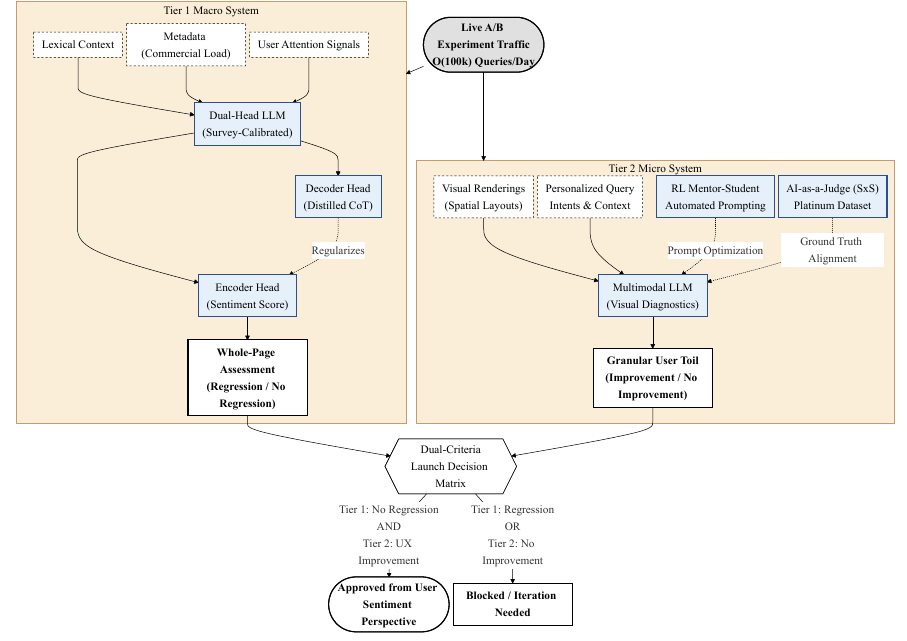}
    \caption{The Dual-Tier Sentiment Framework. Experiments are evaluated in parallel: Tier 1 safeguards against whole-page sentiment regressions, while Tier 2 validates fine-granular improvements in User Toil. A strict launch gate requires both conditions to be met.}
    \label{fig1: architecture}
\end{figure*}

To overcome these limitations and bridge the gap between scale and granularity, we introduce the \textbf{Dual-Tier Sentiment Framework}, an industrial evaluation architecture for offline measurement (Figure \ref{fig1: architecture}). Rather than a simple sequential triage, this framework defines a strict, dual-criteria launch paradigm by running two synergistic systems in parallel:

\textbf{Tier 1 (Macro System)} functions as a global safeguard, providing regression assessment of the user's general sentiment across the whole page. We designed a dual-head LLM based on Gemini XS (a model with $\mathcal{O}$(billions) of parameters) fine-tuned from historical user survey responses. The encoder predicts a user satisfaction score based on a rich suite of lexical, structural, and behavioral telemetry features, while the decoder generates Chain-of-Thought (CoT) {\citep{wei2022chain}} rationales distilled from a larger teacher model. This model ingests lexical context, quantitative metadata, and aggregate user interaction signals to accurately flag broad sentiment regressions.

\textbf{Tier 2 (Micro System)} operates as a fine-granular assessment platform. It utilizes a prompt-based Multimodal Large Language Model (MLLM), specifically Gemini 3.0 Pro, to assess visual renderings of result blocks alongside personalized query contexts. By parsing spatial behavior and visual clarity directly from rendered pages, it actively measures improvements in localized ``User Toil.'' To ensure rigorous evaluation and establish a reliable ground truth, we leverage an AI-as-a-Judge approach for side-by-side (SxS) comparisons, which bootstraps a high-quality human-annotated dataset and actively refines our human rating guidelines. We further enhance Tier 2's robustness in production through a novel Mentor-Student framework optimized with reinforcement learning (RL) for automated prompt generation.

Ultimately, an A/B experiment is only approved for production launch from a user sentiment perspective if it achieves fine-granular UX improvement via the Micro System without whole-page regressions in the Macro System. This dual-requirement is necessary because the two tiers do not always align; the Macro System is less sensitive to granular quality changes, resulting in experiments that yield neutral Tier 1 results despite meaningful metric movements in Tier 2. 

In summary, our main contributions are:
\begin{itemize}
    \item A Dual-Tier Sentiment Framework that redefines launch criteria at scale ($\mathcal{O}(100\text{k})$ queries/day) by requiring both whole-page sentiment stability and granular visual UX improvements.
    \item Deployment of a dual-head, CoT-distilled LLM (Gemini XS Macro System) that reduces measurement latency from weeks to hours and cuts survey overhead by over 40\%.
    \item An MLLM evaluation pipeline leveraging an AI-as-a-Judge for side-by-side (SxS) comparisons, which facilitates the construction of a high-quality human-annotated dataset.
    \item An RL-optimized Mentor-Student prompting strategy for MLLMs (Gemini 3.0 Pro) along with the AI-as-a-Judge.
\end{itemize}

\section{Related Work}

Traditional search result quality evaluation has historically relied on implicit behavioral click models and utility metrics, which often miss the complex visual layouts and cognitive friction of modern web navigation \citep{borisov2018clicksequencemodelweb, shi2025pupillometric}. Consequently, recent academic efforts have transitioned to Multimodal Large Language Models (MLLMs). Architectures such as UGround \citep{gou2024uground} and Aria-UI \citep{pi2025aria} show that grounding raw bounding box coordinates significantly enhances pixel-level spatial understanding \citep{chen2023shikra}. However, these models typically evaluate isolated screenshots in controlled environments and lack mechanisms to capture dynamic pre-click scanning effort or post-click user toil. Moreover, recent benchmarking of MLLMs as UI evaluators \citep{luera2025mllm} reveals that off-the-shelf models align with human judgments on some UI dimensions but break down on others. This validates our Tier 2 design: because base MLLMs struggle with specific UI nuances out-of-the-box, zero-shot inference is insufficient. We overcome this by utilizing RL Mentor-Student tuning and AI Judge alignment to capture complex visual-spatial dimensions where standard MLLMs typically fail.

Industrial high-throughput classifiers often sacrifice interpretability for latency. Recent architectures address this by leveraging large teacher models to distill Chain-of-Thought (CoT) reasoning into lightweight student models \citep{kumar2026enhancing}. Advanced multi-task learning paradigms, such as Dual-Head Reasoning Distillation (DHRD), further utilize this CoT rationale generation solely as a train-time regularizer, boosting accuracy without inference delays \citep{xu2025dualhead}. Additionally, while automated evaluation frameworks like CLEAR efficiently cluster systemic LLM-as-a-judge errors \citep{yehudai2026clear}, and discrete Reinforcement Learning (RL) methods like RLPrompt optimize black-box model instructions \citep{deng-etal-2022-rlprompt}, these tools are typically benchmarked on deterministic reasoning tasks. They lack the cohesive integration needed to resolve the ambiguity and subjectivity of human UX sentiment. Our proposed framework bridges this gap.

\section{System Design and Implementation}

To quantify user satisfaction as described in the dual-criteria launch paradigm introduced in Section 1, our architecture evaluates live experiment traffic through two synergistic pipelines. The Macro System (Tier 1) operates on aggregate telemetry to provide a high-throughput global safeguard against sentiment degradation. Concurrently, the Micro System (Tier 2) leverages multimodal capabilities to diagnose spatial and temporal friction when navigating through web search results.

\subsection{Tier 1: Macro Sentiment Detection}
The Macro System serves as the first line of defense, computing aggregate user satisfaction scores across control and treatment arms to detect whole-page experience degradation with statistical significance. 

\subsubsection{Dual-Head Model Architecture}

Operating as the framework's primary screening mechanism, Tier 1 must support an inference throughput of $\mathcal{O}(1\text{M})$ queries. By rapidly identifying experiments that cause whole-page sentiment degradation, this tier acts as an early-stopping gate—effectively bypassing the need for slow, costly human surveys and avoiding computationally intensive visual analysis in Tier 2. To achieve this scale under strict production latency constraints, we designed a hybrid encoder-decoder architecture that processes only lightweight, text-based modalities. To balance predictive accuracy with interpretability, we separate the tasks across two distinct heads: the encoder functions as a discriminatory classifier predicting binary user sentiment (Regression vs. No Regression), while the decoder is tasked with generating Chain-of-Thought (CoT) rationales.

During the training phase, the reasoning targets are distilled from a massive ``teacher'' model that processes historical human survey responses and rich text features. During live inference, the encoder head directly outputs sentiment predictions while the decoder concurrently synthesizes reasoning. Crucially, the multi-task learning signal from the decoder's CoT generation acts as a structural regularizer. This dual-head paradigm improves the encoder’s predictive Precision-Recall Area Under Curve (PR-AUC) by an absolute margin of +2.0\%.

\subsubsection{Feature Representations}
To capture the nuanced context of search and advertising, the macro model ingests three distinct categories of sparse telemetry:
\begin{itemize}

    \item \textbf{Lexical Context (TEXT)} (Base textual representations, encompassing the search query text, sponsored ad titles, and organic search snippets).
    
    \item \textbf{Quantitative Metadata (QF)} (Query-level structural features, including ad density, commercial load, match types, and high-level UI layout configurations).
    
    \item \textbf{Aggregate User Interactions Features (UF)} (Aggregate behavioral signals reflecting engagement, such as click-through rates, and post-click abandonment).
\end{itemize}

\subsection{Tier 2: Micro Diagnostic System}

\begin{figure}[t]
  \centering
  \includegraphics[width=\columnwidth]{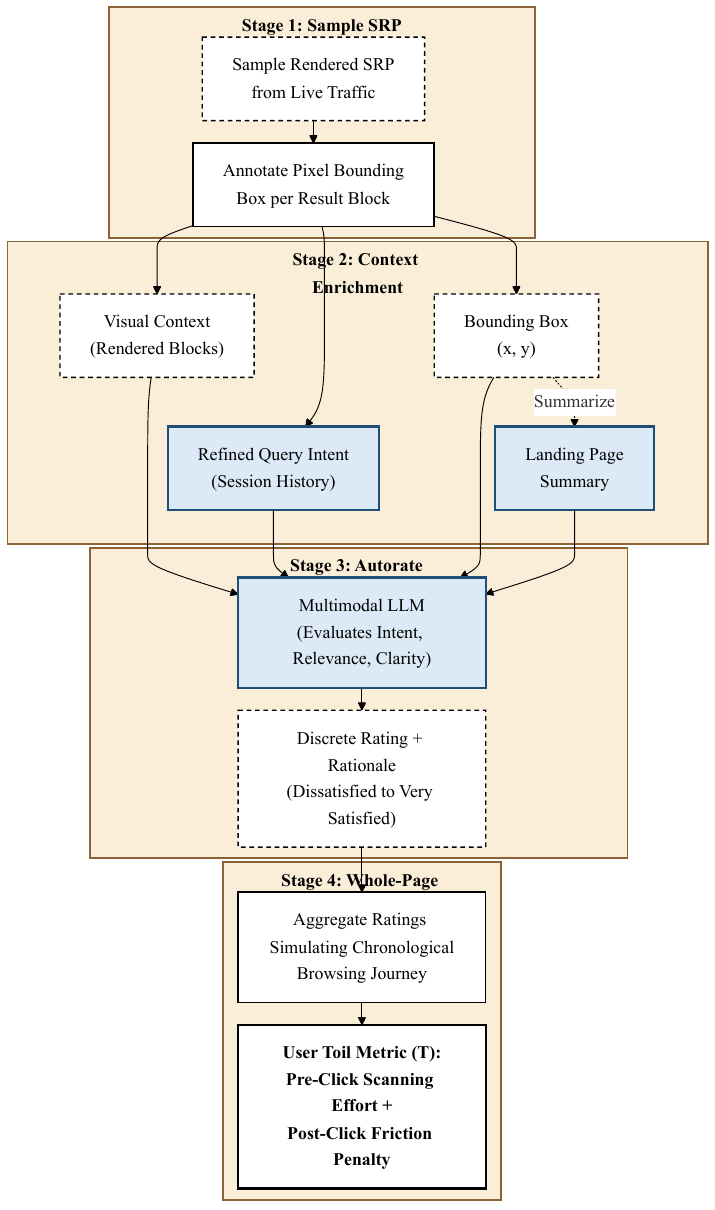}
  \caption{Four-stage diagnostic execution pipeline for computing whole-page User Toil.}
  \label{fig:pipeline}
\end{figure}

While Tier 1 captures telemetry-driven regressions, it lacks the spatial awareness to diagnose \textit{why} a visual UI change frustrated a user and \textit{how} to attribute the root cause on a fine-granular level. The Micro System complements this limitation by quantifying the granular, multimodal friction—termed \textbf{User Toil}—that individuals experience when navigating a Search Results Page (SRP). This is achieved through a structured, four-stage execution pipeline 
(Figure~\ref{fig:pipeline}).

\begin{enumerate}
    \item \textbf{Sample SRP \& Annotate:} The pipeline samples visual renderings from live traffic and annotates exact pixel bounding-box coordinates for each distinct result block.

    \item \textbf{Context Enrichment:} Raw visual renderings are enriched with (a) \textit{Visual Context} via the rendered representation of result blocks; (b) \textit{Refined Query Intent (RQI)} distilled from aggregate past history; and (c) \textit{Result-Specific Context}, a semantic summary of its associated Landing Page (LP).

    \item \textbf{Autorate Satisfaction:} An MLLM ingests the enriched contexts to evaluate intent alignment, semantic relevance, and visual clarity, assigning each block an ordinal satisfaction rating.
    
    \item \textbf{Compute Whole-Page User Toil:} Finally, the pipeline aggregates these discrete ratings into the holistic User Toil metric, simulating the chronological friction of the user's browsing journey.
\end{enumerate}

\subsubsection{Parameterizing the User Toil Metric}
The core philosophy of User Toil is to estimate the total cognitive and temporal effort required for a user to scan a page and interact with results until they reach their first ``Very Satisfied'' (VS) outcome. We define User Toil ($T$) as:

\begin{equation}
\begin{split}
T &= \sum_{i \in \text{Results}} \bigg\{ \Big( t_{\text{scan}}(x_i, y_i) \\
  &\quad + t_{\text{friction}}(LP_i) \Big) \cdot \mathbb{I}(i \le \text{idx}_\text{VS}) \bigg\}
\end{split}
\end{equation}

where $\mathbb{I}$ is an indicator function that truncates the summation once the user encounters the first Highly Satisfactory result ($\text{idx}_{\text{VS}}$), assuming their core search intent has been fulfilled.

The two primary components are parameterized as follows:
\begin{itemize}
    \item $t_{\text{scan}}(x_i, y_i)$ models \textbf{pre-click visual scanning effort}. It is parameterized by the bounding box coordinates $(x_i, y_i)$ of result $i$. Blocks positioned lower on the page (high $y$-coordinate) or buried in dense visual clusters incur a higher $t_{\text{scan}}$ penalty due to the required visual parsing.
    \item $t_{\text{friction}}(LP_i)$ models \textbf{post-click cognitive load}. It represents the estimated time wasted exploring a result. If a clicked landing page ($LP_i$) is irrelevant, clickbait, or difficult to navigate, the MLLM assigns a high $t_{\text{friction}}$ penalty. Conversely, an immediately relevant page incurs minimal friction.
\end{itemize}

To illustrate the practical application of the User Toil metric ($T$), consider a user searching for ``noise-canceling headphones''.

\begin{enumerate}
    \item \textbf{Result 1 (Top Sponsored Product):} Appears at the top of the viewport. The estimated pre-click visual scan cost $t_{\text{scan}}$ is near zero. Upon interaction, the landing page is out-of-stock and difficult to navigate. The MLLM, analyzing the rendered landing page content, rates this outcome as "Dissatisfied," assigning a high post-click friction cost, $t_{\text{friction}} = 15$ units.
    \item \textbf{Result 2 (Organic List):} Positioned lower on the page, requiring an estimated pre-click visual scan cost of $t_{\text{scan}} = 3$ units based on its vertical position and surrounding layout density. The landing page content has the exact desired product. The MLLM rates this as "Very Satisfied," with $t_{\text{friction}} = 0$.
\end{enumerate}

The total estimated User Toil for this page is $(0 + 15) + (3 + 0) = 18$. If a UI experiment moved Result 2 to the top, a user would likely interact with it first. The new estimated Toil would drop to $(0 + 0) = 0$, quantitatively demonstrating a significant UX improvement. 

The costs per result are model-based estimates derived from the visual rendering and content, not direct measurements of individual user behavior. \rev{We do not use raw per-session dwell times as ground-truth, since they introduce severe heavy-tailed variance. For example, idle browser tabs or extended reading on informational pages conflate benign engagement with UX friction and inflate experiment confidence intervals.} Specifically, these units act as ordinal penalty scores weighted by historical timing distributions. Through an offline calibration mapping, the discrete categorical outputs of the MLLM (e.g., ``Dissatisfied'') are mapped to numerical penalty weights that reflect the relative magnitude of time-wasting observed in historical log distributions. Consequently, a reduction in User Toil serves as an ordinal threshold for our launch gate rather than a literal budget of measured seconds. In offline stability checks, directional experiment-level Toil movements and Tier 2 launch recommendations remain stable across alternative aggregations of historical penalty weights that preserve ordinal monotonicity.

\subsection{Iterative Alignment and Prompt Optimization}
To ensure the Micro System accurately reflects complex human subjectivity, we designed an automated alignment architecture that continuously refines the MLLM's behavior without requiring full-parameter fine-tuning.

\subsubsection{AI-as-a-Judge for Loss Pattern Analysis}

We implemented an auxiliary MLLM (Gemini 3.0 Pro) equipped with toolkits for Google Search and URL context parsing to act as an ``AI Judge'' for Side-by-Side (SxS) evaluation. By ingesting human rating guidelines alongside the outputs of both human experts and our autorater, the AI Judge systematically clusters failure patterns. For instance, the Judge identified that early iterations of our baseline struggled to distinguish boundaries in complex carousels (e.g. UI elements that group multiple individual result cards horizontally). This specific diagnostic insight drove the architectural decision to inject exact pixel bounding boxes into the prompt, yielding an immediate +5\% F1 improvement against human ground truth.

\subsubsection{RL-Based Prompt Optimization}


\begin{figure*}[t]
    \centering
    \includegraphics[width=0.85\textwidth]{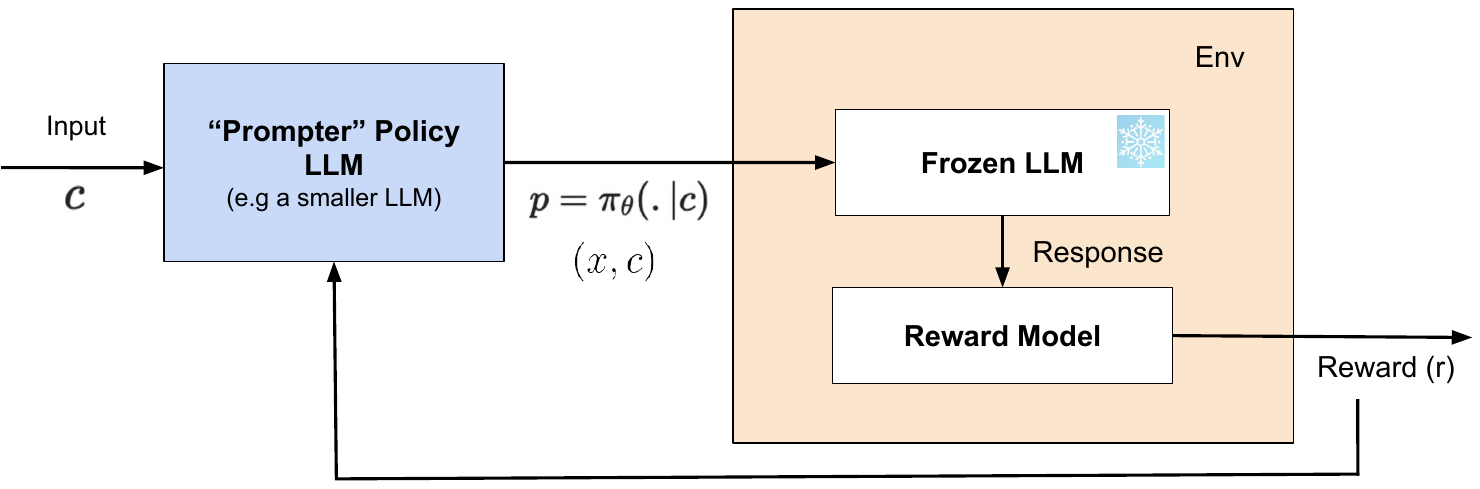}
    \caption{Dual-agent RL-based Automated Prompt Optimization flow.}
    \label{fig:apo_rl}
\end{figure*}

To scale this alignment, we deployed a dual-agent Reinforcement Learning (RL) framework \citep{sayana2026prompting} to automate prompt construction (Figure~\ref{fig:apo_rl}). 

\textbf{Dual-Agent Dynamics:} The architecture consists of a \textit{Mentor} (a lightweight, highly efficient model) and a \textit{Student} (a massive, frozen reasoning engine). The Mentor does not evaluate sentiment; instead, it functions as a strategic controller, learning a policy ($\pi_{\theta}$) to formulate optimal, task-specific system instructions ($p$) token-by-token. The Student acts as the worker agent, executing these instructions to rate the SRP. Standardizing execution on a frozen Student ensures the backbone retains its broad multimodal capabilities without risking catastrophic forgetting.

\textbf{Optimization Regime:} We formulate the optimization of the Mentor policy $\pi_{\theta}$ as an RL problem over a discrete token action space:

\begin{equation}
\max_{\theta} \, \mathbb{E}_{x \sim \mathcal{D}, p \sim \pi_{\theta}(\cdot|x)} \Big[ R(x, p, y) - \beta \mathbb{D}_{KL} \left( \pi_{\theta} \| \pi_{\text{ref}} \right) \Big]
\end{equation}

where $x$ is the input context, $y$ is the ground truth human rating, $p$ is the generated prompt, $\mathbb{D}_{KL}$ is the Kullback-Leibler divergence to prevent policy collapse, and $\beta$ is the regularization coefficient. 

A unique challenge in UX evaluation is optimizing a single query-level prompt ($p$) that must accurately evaluate multiple, distinct results on the same page. To preserve this structural hierarchy, we design the reward $R$ as the arithmetic mean of the individual result rewards $r_i$ (where $r_i$ reflects the MLLM's accuracy against the human rating for result $i$): $R = \frac{1}{n} \sum_{i=1}^{n} r_i$. It ensures that the Mentor is uniformly incentivized to generate instructions that improve the Student's diagnostic accuracy across the entire web page, effectively bridging the gap between automated telemetry and subjective human experience.

\rev{The RL training is performed entirely offline to remove prompt-generation variance in production: once $\pi_\theta$ converges, the single highest-scoring discrete system prompt $p^*$ (selected on a held-out validation split) is frozen and deployed to the Student MLLM. The student MLLM also utilizes deterministic greedy decoding configurations ($\text{e.g. temperature}=0$) to remove sampling variance in its outputs during production A/B evaluation.}

\section{Experimental Methodology and Results}

To validate the Dual-Tier Sentiment Framework, we designed an evaluation methodology that mirrors its production deployment: Tier 1 is evaluated on its ability to act as a high-throughput, high-recall regression net, while Tier 2 is evaluated on its diagnostic precision against human-perceived visual friction.

\subsection{Tier 1: Macro User Sentiment Detection}

\subsubsection{Dataset: Survey-Backed Telemetry}
Evaluating subjective user friction at a whole-page scope requires proprietary datasets that accurately reflect real-world UI complexity. Ground truth labels for Tier 1 were derived from in-product user surveys triggered after a few minutes of active search session time. 

To ensure robust generalization, the Macro model was trained on data from 11 distinct A/B experiments. It was rigorously evaluated on a holdout suite of 12 unseen experiments collected strictly after the model checkpoint was frozen. This evaluation dataset comprises $>500,000$ records sampled evenly across control and treatment arms to calculate experiment-level precision and recall.

\subsubsection{Ablation and Impact}

\begin{table}[t]
\centering
\small
\begin{tabularx}{\columnwidth}{l X c}
\hline
\textbf{Model} & \textbf{Architecture / Features} & \textbf{$\Delta$ PR-AUC} \\
\hline
Baseline & Lexical features (TEXT) & (Ref) \\
+ Decoder & TEXT + Teacher CoT & +2.0\% \\
+ Metadata & TEXT + Query Features (QF) & +5.1\% \\
+ User Sigs & TEXT + QF + User Features (UF) & \textbf{+9.8\%} \\
\hline
\end{tabularx}
\caption{Ablation Study on Macro System Architecture.}
\label{tab:ablation}
\end{table}

We fine-tuned the Macro Model using PR-AUC as our primary metric to account for class imbalances in sentiment degradation. Extensive ablation studies (Table~\ref{tab:ablation}) quantify our architectural decisions. The baseline model, utilizing solely lexical features (TEXT), served as the reference and represents the prior production incumbent deployed on the same training data. Introducing the dual-head CoT regularizer brought an immediate +2.0\% PR-AUC gain. The sequential integration of Quantitative Metadata (QF) and Aggregated User Interactions Features (UF) signals pushed the peak performance gain to +9.8\% over the baseline.

\textbf{Translating Metrics to Operational Impact:} This +9.8\% PR-AUC gain fundamentally alters the A/B testing lifecycle: We define a sentiment regression as a statistically significant degradation in the aggregated user satisfaction metric (treatment versus control), verified using a required volume of events. Under this definition, Tier 1 successfully yields no missed regressions among the 4 validated positive regressions across the 12 held-out experiments. Furthermore, precision is evaluated at the experiment level: at the operational gating threshold, the baseline model flagged 6 experiments (yielding a precision of 66.7\%, i.e., 4/6), while the final dual-head Macro System flagged 5 experiments, yielding a precision of 80\% (4/5)—an effective $\sim$20\% relative detection precision improvement. As a reliable early-stopping gate, it intercepts degrading experiments immediately, reducing measurement latency from weeks to hours, and cutting human survey overhead costs by over 40\% in production.

\subsection{Tier 2: Micro Diagnostic System}

While Tier 1 is evaluated on $\mathcal{O}(100\text{k})$ labels, Tier 2 relies on highly precise, multimodal alignment. A common challenge in MLLM evaluation is the scarcity of fine-grained spatial ground truth. To overcome this, we engineered two distinct datasets and leveraged our AI Judge to enforce label quality.

\subsubsection{Bootstrapping the Human-annotated Dataset via AI Judge}
Rather than blindly training a model on a massive, noisy dataset with high overfitting risk, our paradigm relies on a \textit{frozen} foundational MLLM whose instructions are aligned via prompting. In this paradigm, label fidelity outweighs label volume.

To establish this fidelity, we extracted result blocks from live web renderings and tasked independent human raters with  detailed editorial guidelines. \rev{Each extracted result block is double-annotated by independent raters on a four-tier ordinal satisfaction scale followed by expert adjudication to resolve ties. Because Equation~(1) truncates toil accumulation at $\text{idx}_{\text{VS}}$, evaluation operationalizes binary \textit{Very Satisfied} vs.\ \textit{Rest} classification on a roughly balanced distribution.} We deployed an AI Judge to systematically compare conflicting human labels to investigate the initial high inter-rater disagreement rate. This automated loss analysis exposed core ambiguities in how humans perceived boundaries in complex UI carousels. By iteratively refining the evaluation guidelines based on the AI Judge's insights, we substantially increased dual-rater agreement across calibration rounds and stabilized the human rating process. \rev{Because each query encompasses multiple evaluated result blocks across organic and sponsored slots, sampling $O(100)$ queries for prompt optimization and a strictly isolated post-freeze holdout split of $O(100)$ queries yields several hundred adjudicated result-level labels per split. This process provides over 1,000 expert-validated result labels in total when combined with earlier guideline-calibration batches.}

\subsubsection{Validation against First-Result Surveys}
To prove that our expert labels accurately map to genuine user perception at scale, we deployed a novel ``First-Result'' user survey. Users were presented with the first viewport of a live SRP and asked to rate and justify their friction for only that specific result. To mitigate the inherent noise of human surveys, UX researchers and product managers established strict guidelines to filter out objective spam such as nonsensical text or users evaluating the wrong UI element, yielding >5000 high-quality, real-user labels. Crucially, this holdout survey pool was kept completely isolated from the AI Judge. To ensure the evaluation strictly measures visual-spatial UI friction (User Toil), we denoised this pool by filtering out two specific categories of responses: (1) Invalid execution of the task (e.g., junk responses, evaluating the wrong UI element, or admitted misclicks); and (2) Out-of-scope platform biases (e.g., systemic AI aversion or subjective baseline differences that do not reflect UI layout quality). The final denoised pool retained >60\% of the initial labels, ensuring the reported performance metrics evaluate our specific operational target rather than the raw, unfiltered survey distribution. \rev{To prevent data cleaning from introducing confirmation bias in favor of the autorater, the denoising classifier inspects only the user's open-ended element description, rationale, and web result context. The survey cleaning process remains strictly blind to Tier~2's prompt or predicted rating.}

\subsubsection{Alignment Results}
The Micro System's baseline prompt was initially optimized using the AI Judge's win/loss pattern analysis. Injecting explicit pixel-coordinate features and applying Human-in-the-Loop prompt refinements yielded a $+5\%$ F1 improvement \rev{by resolving salient structural errors such as carousel boundary bleed. Beyond this plateau, further manual prompt edits exhibited precision-recall oscillation---tightening rubric constraints for specific failure modes improved precision on targeted slices while regressing recall overall}. Transitioning to the automated Reinforcement Learning (RL) framework overcame this trade-off, further improving global F1 by $>1.5\%$. Crucially, when evaluating the frozen RL-optimized prompt policy against the completely independent pool of $O(1000)$ denoised real-user survey responses, the model sustained this F1 gain, empirically proving that our prompt-aligned Micro System generalizes across both expert-adjudicated and real-user ground truth at an industrial scale.

\section{Deployment Challenges and System Architecture}

Deploying a multimodal evaluation framework at $\mathcal{O}(100\text{k})$ daily queries introduces unique infrastructural complexities.

\textbf{Bridging the Infrastructure Gap}:
A primary bottleneck was the Visual Rendering Dependency. Initial deployment of Tier 2 was blocked because the efficient infrastructure to render web results for model inference purposes did not exist. To resolve this, we engineered an asynchronous result-rendering pipeline. By heavily parallelizing the log-join and web-content-replay stages and leveraging deployment cell load balancing, we reduced the data processing time for a full A/B experiment cohort from several days to a few hours. 

\textbf{Inference Cost and Latency Bounds}:
Deployed for offline evaluation, the framework avoids strict real-time latency constraints but requires massive sustained throughput. The computational cost of Tier 2 prevents synchronous, real-time serving. Generating bounding boxes, summarizing landing pages, and running inference through a massive MLLM (with a parameter count orders of magnitude larger than our macro model) for every query is economically prohibitive. Therefore, Tier 1's lightweight encoder-decoder architecture operating at a scale of $\mathcal{O}(\text{billions})$ parameters is necessary to handle the raw scale. The expensive Tier 2 pipeline is invoked only when fine-granular attribution is required. \rev{This asynchronous batch design aligns with its operational role given that A/B launches inherently require accumulating hours of live experiment impressions to achieve statistical power. Therefore, a multi-hour batch turnaround introduces no bottleneck to experiment launch decisions.}

\section{Lessons Learned \& Best Practices}

Our journey from developing a telemetry-based screening tool to a granular, multimodal diagnostic engine yielded critical insights into user behavior and LLM deployment.

\paragraph{Isolating UX Friction from User Bias:} We discovered raw ``negative sentiment'' in surveys is frequently skewed by factors unrelated to content quality (e.g., inherent ad or AI content aversion). Rigorous data-cleaning and AI Judge adjudication were essential to isolate genuine spatial-UI friction (User Toil) from platform-level bias.

\paragraph{Stability of Predicted Magnitude:} We attempted to use generative LLMs for Tier 1 initially. While directionally predictive, their predicted \textit{magnitude} lacked numerical stability to consistently align with historical surveys. Consequently, we adopted the fine-tuned encoder-decoder for reliable regression screening.

\paragraph{The Necessity of Contextual Intent Alignment:}
Early Micro System iterations often neglected query intent context, rating results as ``Satisfied'' based purely on visual aesthetics while disregarding deeper search goals of the user. Forcing the model to ingest Refined Query Intent (RQI) signals derived from aggregated history data was critical in shifting the model's focus from mere visual quality to true cognitive friction.

\paragraph{Automated Optimization vs. Manual Prompt Tuning:}
Finally, we highlight the severe limitations of manual prompt engineering in multimodal UX evaluation. Human engineers naturally over-index on recent, salient failure cases, often regressing overall accuracy when attempting to patch edge cases. By abstracting prompt construction to a dual-agent RL framework and auto prompt adjustment from win/loss patterns, we not only achieved higher F1 score but also eliminated the human engineering overhead required to maintain the system as search UI layouts continuously evolve.

\section{Limitations}

While our framework advances automated UX attribution, several limitations remain. First, the computational cost of multimodal rendering and MLLM inference restricts Tier~2 to asynchronous batch evaluation for A/B launch gating rather than synchronous real-time serving. Second, the empirical benchmarks reported in this work focus on US-based mobile traffic; \rev{although the underlying multilingual backbones and 2D bounding-box indexing support broader deployment, full validation across multi-column desktop layouts and non-US locales requires device- and region-stratified $t_{\text{scan}}$ recalibration and localized rater guidelines}. Third, \rev{calibrating and validating against user surveys requires LLM-assisted denoising to isolate layout toil from platform priors; although blind multi-pass consensus mitigates confirmation bias, residual sensitivity to filtering criteria remains}. Finally, our current production evaluation scores static page renderings; \rev{extending attribution to interactive UI states (e.g., carousels, dynamic filters, and video previews) via interaction-conditioned multi-state rendering remains an active direction of ongoing work}.

\section{Ethical Statement}
The development and deployment of the Dual-Tier Sentiment Framework involve processing aggregate user interaction data and visual representations of search results, necessitating careful ethical considerations:
\begin{enumerate}
\item Data Privacy and Anonymization: All telemetry data used for training and evaluating both Macro and Micro systems is anonymized. Privacy is strictly protected through tiered access controls and rigorous organizational policies.

\item Purpose Limitation: The data and models are used exclusively for evaluating and improving search and ads quality and user experience. The analysis focuses on identifying UX frictions such as visual clutter or content quality issues.

\item Data Retention: The collected data and associated analyses are subject to a strict time-to-live data retention policy, in line with internal data handling policies.
\end{enumerate}

By combining rigorous data protection and privacy with user sentiment modeling, we ensure our web results deliver genuine value and meet real information needs.

\bibliography{aaai2027}


\end{document}